\documentclass[dvipdfmx]{article}
\usepackage{arxiv}
\usepackage{graphicx}
\usepackage{subfigure}
\usepackage{algpseudocode}
\usepackage{algorithm}
\usepackage{amsmath}
\usepackage{newtxtext,newtxmath}
\usepackage{color}
\usepackage{bm}
\usepackage{url}
\usepackage{enumitem}
\usepackage{multicol}

\title{Skip2-LoRA: A Lightweight On-device DNN Fine-tuning Method for
  Low-cost Edge Devices}

\author{
  Hiroki Matsutani\\
  Keio University\\
  3-14-1 Hiyoshi, Kohoku-ku, Yokohama, Japan\\
  \texttt{matutani@arc.ics.keio.ac.jp} \\
  \And
  Masaaki Kondo\\
  Keio University\\
  3-14-1 Hiyoshi, Kohoku-ku, Yokohama, Japan\\
  \texttt{kondo@acsl.ics.keio.ac.jp} \\
  \And
  Kazuki Sunaga\\
  Keio University\\
  3-14-1 Hiyoshi, Kohoku-ku, Yokohama, Japan\\
  \texttt{sunaga@arc.ics.keio.ac.jp} \\
  \And
  Radu Marculescu\\
  The University of Texas at Austin\\
  2501 Speedway, Austin, Texas, USA\\
  \texttt{radum@utexas.edu} \\
}

\begin{document}

\maketitle
\begin{abstract}
This paper proposes Skip2-LoRA as a lightweight fine-tuning method
for deep neural networks to address the gap between pre-trained
and deployed models.
In our approach, trainable LoRA (low-rank adaptation) adapters are
inserted between the last layer and every other layer to enhance the
network expressive power while keeping the backward computation cost low.
This architecture is well-suited to cache intermediate computation results of
the forward pass and then can skip the forward computation of seen
samples as training epochs progress.
We implemented the combination of the proposed architecture and cache,
denoted as Skip2-LoRA, and tested it on a \$15 single board computer.
Our results show that Skip2-LoRA reduces the fine-tuning time by
90.0\% on average compared to the counterpart that has the same number of
trainable parameters while preserving the accuracy, while taking only
a few seconds on the microcontroller board.
\end{abstract}

\section{Introduction}\label{sec:intro}

%
%
On-device learning is an emerging research direction in edge AI aiming
to reduce the gap between pre-trained and deployed models.
Since the available compute resources are limited in edge environments,
full retraining of deep models is hardly feasible; thus, lightweight
retraining methods of neural networks have been studied recently
\cite{Zhu23,Nadalini23,Sunaga23}.

%
%
Such on-device learning methods can be broadly classified into 1) ELM
(extreme learning machine) based retraining and 2) fine-tuning of some
specific layers using a backpropagation algorithm.
In the ELM-based on-device learning \cite{Tsukada20,Sunaga23}, the OS-ELM
(online sequential ELM) algorithm \cite {Liang06} is used for training
the weight parameters of neural networks that have a single hidden layer.
Thus, the ELM-based approach cannot be applied to DNNs (deep neural
networks) that have multiple or many hidden layers; instead, the
backpropagation-based approach can be used for such DNNs.
A well-known method based on backpropagation fine-tunes the last layer
of DNNs \cite{Ren21};
in this case, the backward compute cost is very small compared to the
full training, but the network expressive power remains limited since
only the last-layer weights can be updated.
Another method freezes the weight parameters while only updating the
bias modules \cite{Cai20}.
TinyTL introduces the lite residual module as a generalized bias
module to be fine-tuned \cite{Cai20}.
All these methods update parts of the pre-train model.
%
%
In addition, fine-tuning methods have been widely studied in the
context of LLMs (large language models).
A popular approach in LLMs is to add trainable adapters to pre-trained
networks.
Trainable adapter layers can be inserted in series to pre-trained
networks \cite{Houlsby19}, or attached to pre-trained weight matrixes
in parallel \cite{EHu21}.
LoRA (low-rank adaptation) \cite{EHu21} employs the latter approach;
that is, trainable rank decomposition matrixes are attached to each
layer of a Transformer architecture \cite{Vaswani17}.
This approach is portable, meaning that only the adapters are updated
while the original weights are untouched.

%
%
In this paper, we extend the LoRA-based fine-tuning methods for
resource-limited edge devices.
Starting from adding LoRA adapters to each layer of DNNs, we propose
a lightweight fine-tuning approach called Skip2-LoRA.
Our contributions are summarized as follows:
\begin{itemize}
\item We propose a new architecture where trainable LoRA adapters are
  inserted between the last layer and every other layer to enhance the
  network expressive power while keeping the
  backward compute cost low.
\item This new architecture enables us to cache the intermediate
  compute results during the forward pass and thus skip the forward
  computation of seen samples as training epochs progress.
\item Our experimental results show that Skip2-LoRA reduces the
  fine-tuning time by 89.0\% to 92.0\% compared to the baseline while
  achieving comparable accuracies, while taking only a few seconds on
  a \$15 single board computer.
\end{itemize}

%
%
This paper is organized as follows.
Sections \ref{sec:prelim} and \ref{sec:related} review preliminaries
and basic knowledge of fine-tuning methods.
Section \ref{sec:design} proposes Skip2-LoRA, and Section
\ref{sec:eval} evaluates it in terms of accuracy, execution time, and
power consumption.
Section \ref{sec:conc} summarizes our contributions. 


\section{Preliminaries}\label{sec:prelim}

%
%
Let $N$ and $M$ be the input and output dimensions of an FC 
(fully-connected) layer, respectively, in a DNN.
A forward pass of the FC layer is computed as follows:
\begin{equation}
  \bm{y} = G(\bm{x} \cdot \bm{W} + \bm{b}), \label{eq:fc}
\end{equation}
where $\bm{x} \in \mathbb{R}^{B \times N}$,
$\bm{y} \in \mathbb{R}^{B \times M}$,
$\bm{W} \in \mathbb{R}^{N \times M}$, 
$\bm{b} \in \mathbb{R}^{M}$, and $B$ represent the input feature map,
output feature map, weight parameters, bias parameters, and batch
size, respectively.

%
%
A backward pass of the FC layer is computed as follows:
\begin{eqnarray}
  \bm{gW} &=& \bm{x^\top} \cdot \bm{gy} \\
  \bm{gb} &=& \sum^B \bm{gy} \\
  \bm{gx} &=& \bm{gy} \cdot \bm{W^\top},
\end{eqnarray}
where $\bm{gx} \in \mathbb{R}^{B \times N}$,
$\bm{gy} \in \mathbb{R}^{B \times M}$,
$\bm{gW} \in \mathbb{R}^{N \times M}$, and
$\bm{gb} \in \mathbb{R}^{M}$ represent the gradients
of $\bm{x}$, $\bm{y}$, $\bm{W}$, and $\bm{b}$, respectively.

%
%
$\bm{W}$ and $\bm{b}$ are updated as follows:
\begin{eqnarray}
  \bm{W} &\leftarrow& \bm{W} -\eta \cdot \bm{gW} \\
  \bm{b} &\leftarrow& \bm{b} -\eta \cdot \bm{gb},
\end{eqnarray}
where $\eta$ represents a learning rate.

%
%
A forward pass of a LoRA adapter of rank $R$ for the FC layer is 
computed as follows:
\begin{eqnarray}
  \bm{y_A} &=& \bm{x} \cdot \bm{W_A} \\
  \bm{y_B} &=& \bm{y_A} \cdot \bm{W_B} \label{eq:lora_yb} \\
  \bm{y} &\leftarrow& \bm{y} + \bm{y_B}, \label{eq:lora_y}
\end{eqnarray}
where $\bm{y_A} \in \mathbb{R}^{B \times R}$ and
$\bm{y_B} \in \mathbb{R}^{B \times M}$ are intermediate outputs
to update $\bm{y}$.
$\bm{W_A} \in \mathbb{R}^{N \times R}$ and
$\bm{W_B} \in \mathbb{R}^{R \times M}$ are the weight parameters of
the adapter.

%
%
A backward pass of the LoRA adapter is computed as follows:
\begin{eqnarray}
  \bm{gW_B} &=& \bm{y_A^\top} \cdot \bm{gy} \\
  \bm{gx_B} &=& \bm{gy} \cdot \bm{W_B^\top} \\
  \bm{gW_A} &=& \bm{x^\top} \cdot \bm{gx_B} \\
  \bm{gx_A} &=& \bm{gx_B} \cdot \bm{W_A^\top} \label{eq:lora_gxa} \\
  \bm{gx} &\leftarrow& \bm{gx} + \bm{gx_A}, \label{eq:lora_gx}
\end{eqnarray}
where $\bm{gx_B} \in \mathbb{R}^{B \times R}$ and
$\bm{gx_A} \in \mathbb{R}^{B \times N}$ are intermediate gradients.
$\bm{gW_B} \in \mathbb{R}^{R \times M}$ and 
$\bm{gW_A} \in \mathbb{R}^{N \times R}$ represent the gradients
of $\bm{W_B}$ and $\bm{W_A}$, respectively.

%
%
$\bm{W_A}$ and $\bm{W_B}$ are updated as follows:
\begin{eqnarray}
  \bm{W_A} &\leftarrow& \bm{W_A} -\eta \cdot \bm{gW_A} \\
  \bm{W_B} &\leftarrow& \bm{W_B} -\eta \cdot \bm{gW_B}.
\end{eqnarray}


\section{Baseline Fine-tuning Methods}\label{sec:related}

\begin{figure*}[t]
\begin{minipage}[t]{0.163\linewidth}
	\centering
	\subfigure[FT-All] {
	\includegraphics[height=30mm]{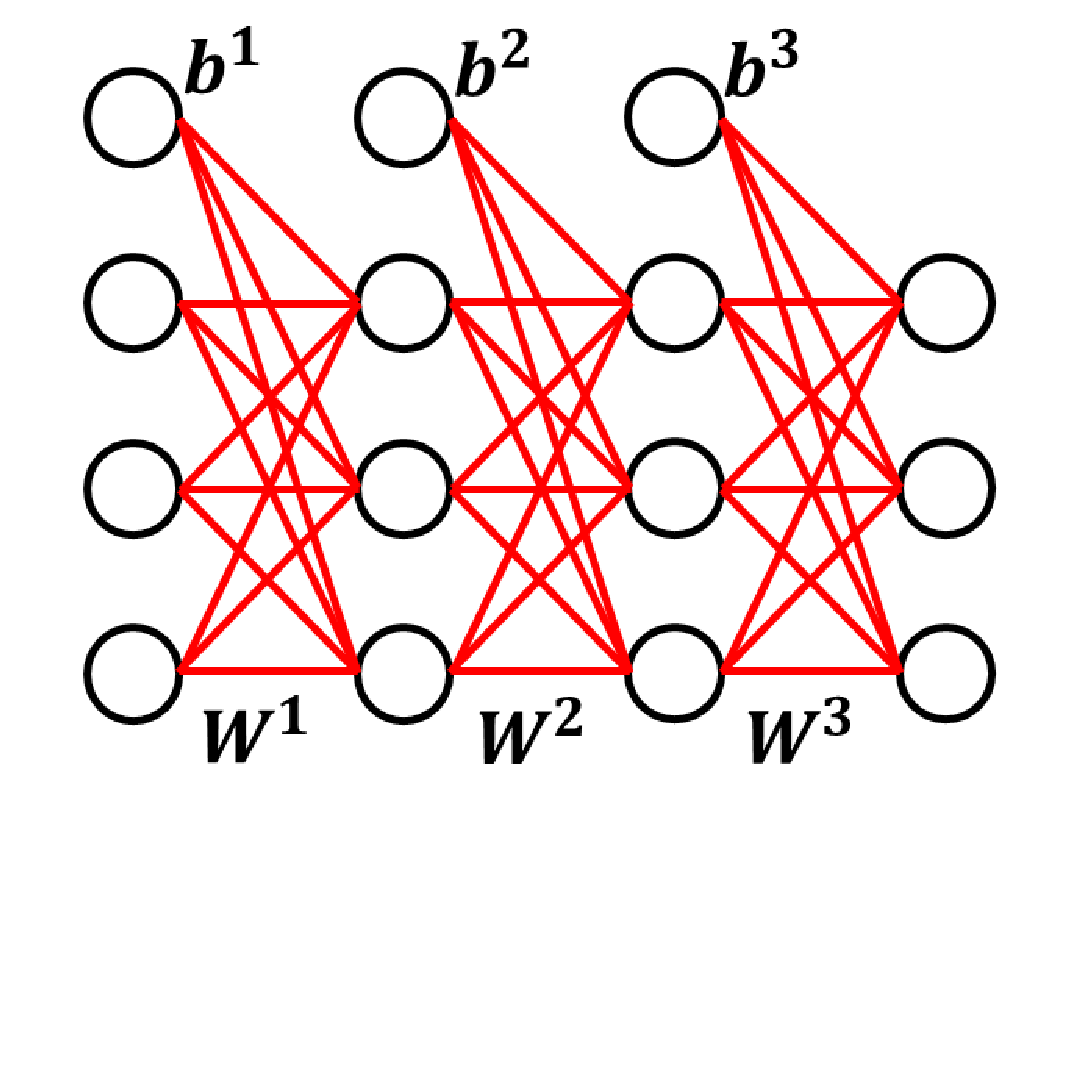}
	\label{fig:ft-all}
	}
\end{minipage}
\begin{minipage}[t]{0.163\linewidth}
	\centering
	\subfigure[FT-Last] {
	\includegraphics[height=30mm]{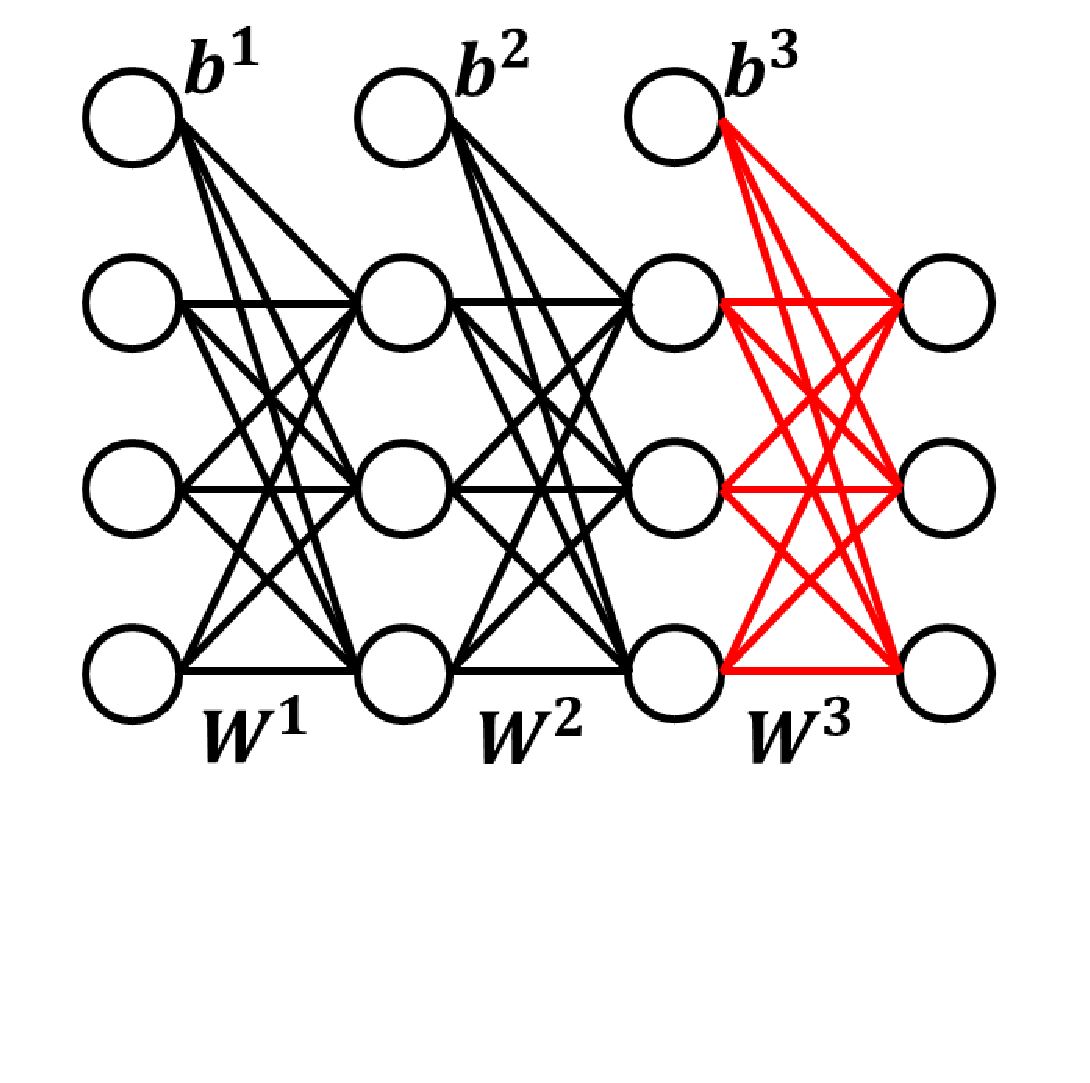}
	\label{fig:ft-last}
	}
\end{minipage}
\begin{minipage}[t]{0.163\linewidth}
	\centering
	\subfigure[FT-Bias] {
	\includegraphics[height=30mm]{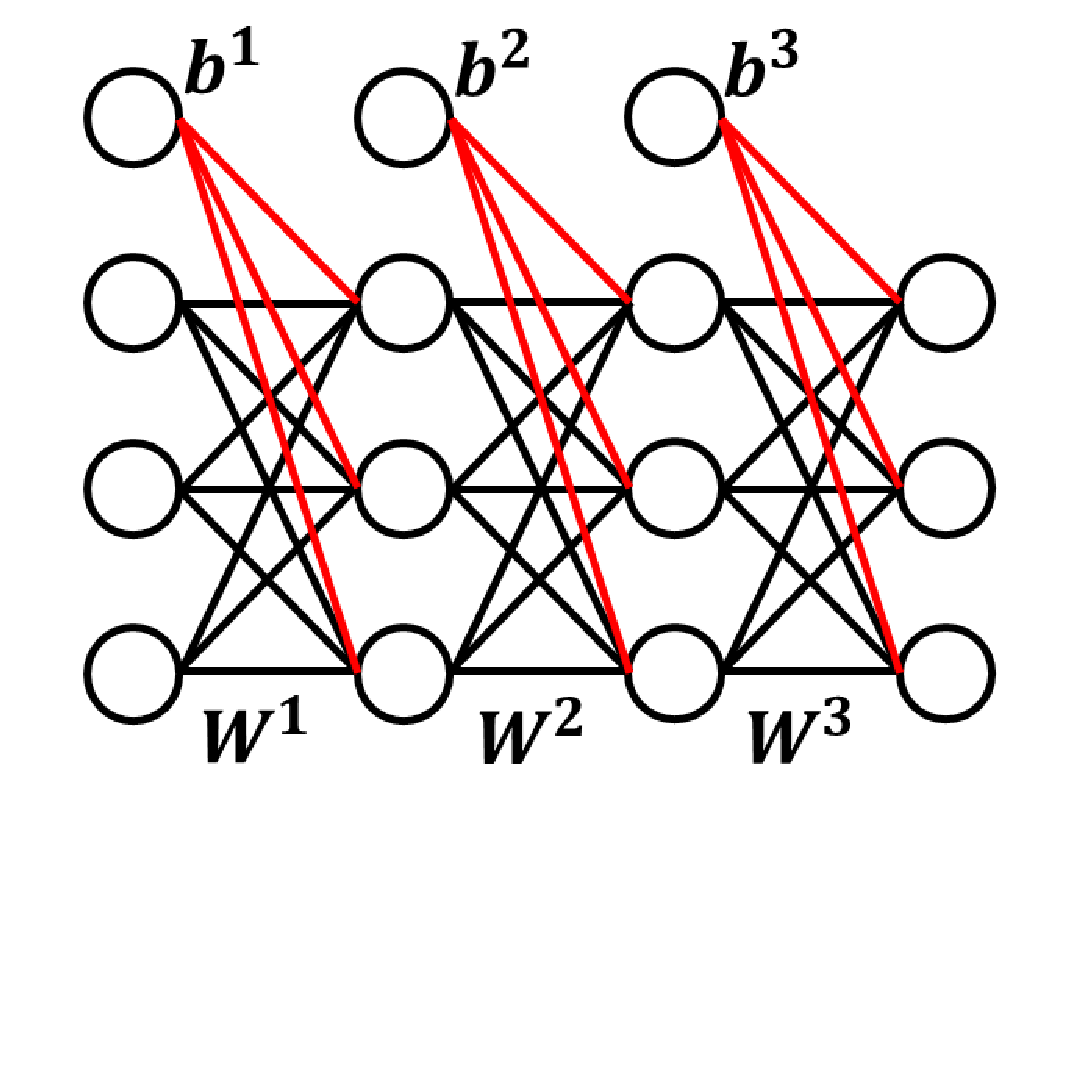}
	\label{fig:ft-bias}
	}
\end{minipage}
\begin{minipage}[t]{0.163\linewidth}
	\centering
	\subfigure[LoRA-All] {
	\includegraphics[height=30mm]{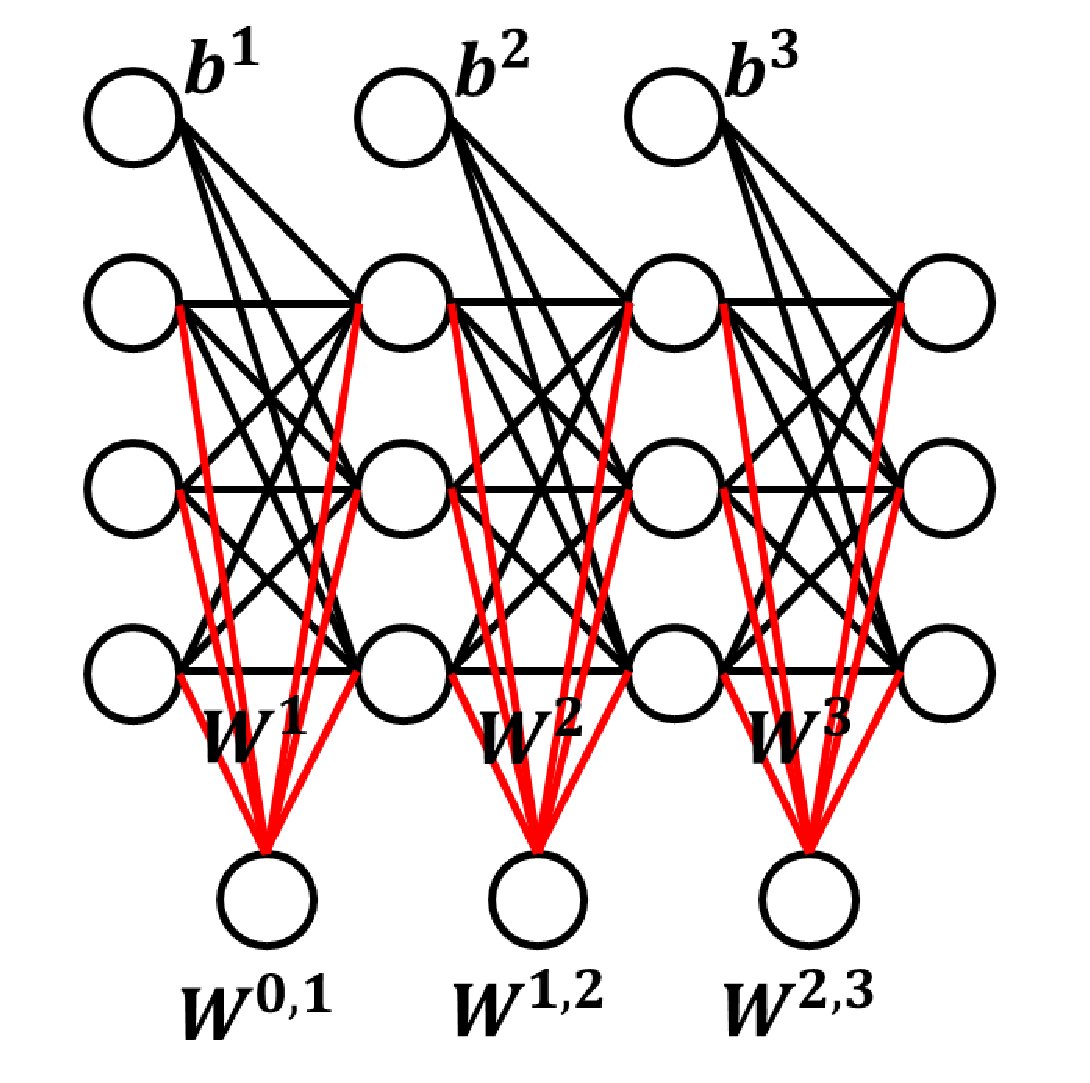}
	\label{fig:lora-all}
	}
\end{minipage}
\begin{minipage}[t]{0.163\linewidth}
	\centering
	\subfigure[LoRA-Last] {
	\includegraphics[height=30mm]{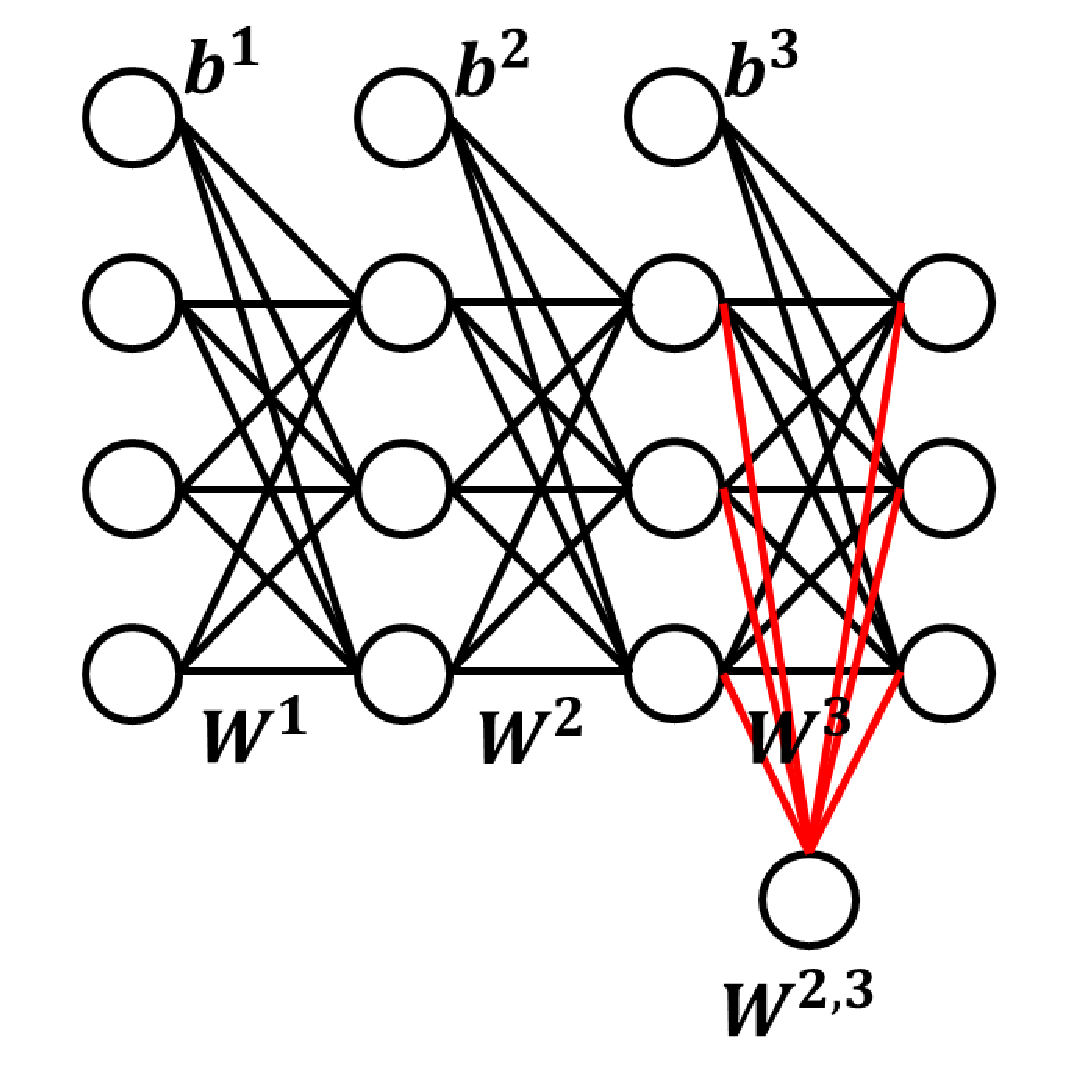}
	\label{fig:lora-last}
	}
\end{minipage}
\begin{minipage}[t]{0.163\linewidth}
	\centering
	\subfigure[Skip-LoRA] {
	\includegraphics[height=30mm]{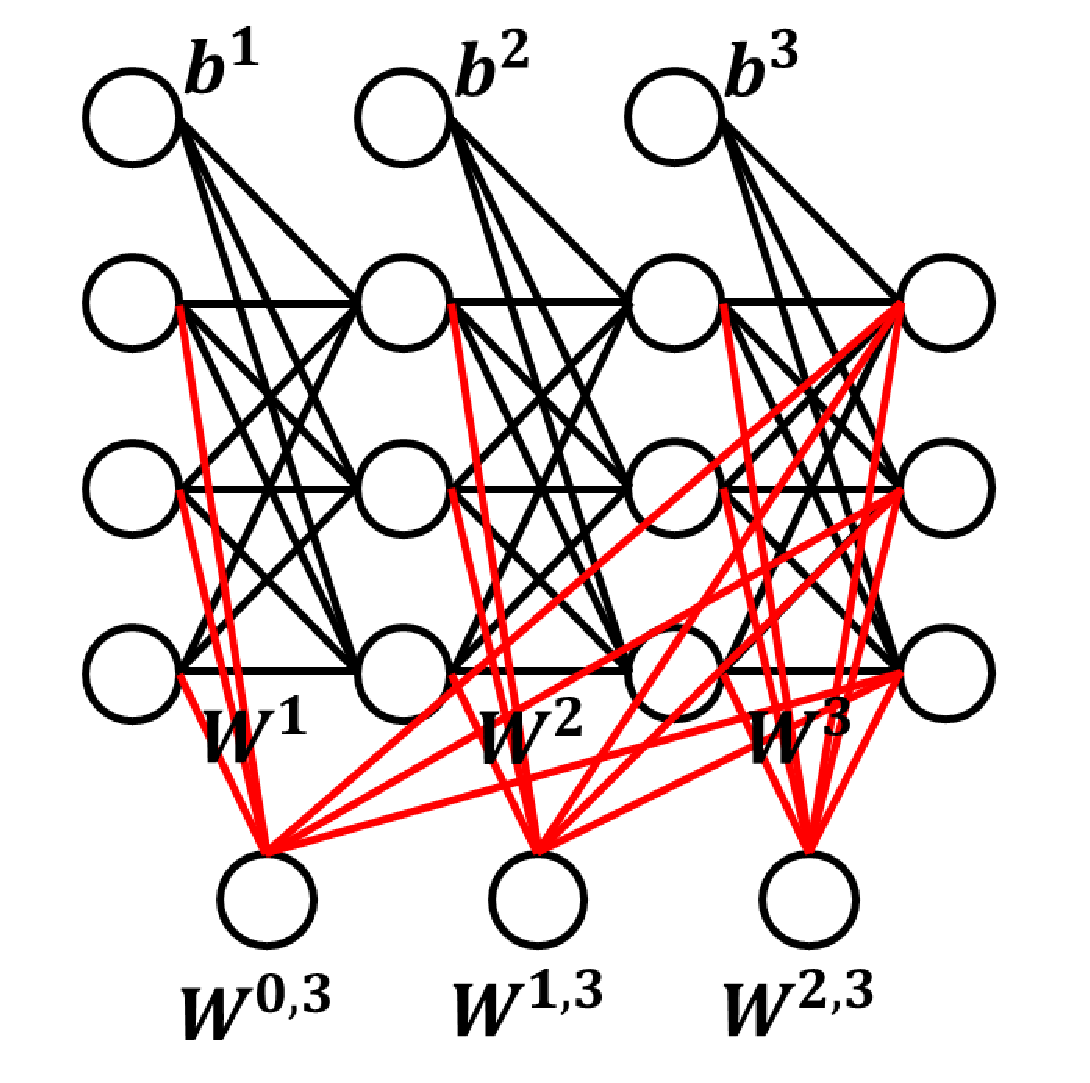}
	\label{fig:skip-lora}
	}
\end{minipage}
\caption{Fine-tuning methods of DNNs consisting of $n$ FC layers, where $n=3$.
  $\bm{W^k}$ and $\bm{b^k}$ denote weights and biases for $k$-th layer.
  In LoRA-All and LoRA-Last, $\bm{W^{k-1,k}}$ denotes weights for
  $k$-th LoRA adapter, where rank $R=1$.
  Parameters to be updated are colored in red.}
\label{fig:ft-method}
\end{figure*}

%
%
A forward pass of an FC layer computes $\bm{y}$, while the backward pass 
computes $\bm{gW}$, $\bm{gb}$, and $\bm{gx}$.
In fine-tuning scenarios, not all are necessary; for example,
$\bm{gW}$ and $\bm{gb}$ of an FC layer are not necessary when weight 
and bias parameters of the layer are not updated.
Compute types of FC layers are classified as listed in the upper half 
of Table \ref{tab:type}.
%
%
The number of floating-point operations and memory size can be modeled
for each compute type, but they are omitted due to the page limitation.

\begin{table}[t]
\centering
\caption{Compute types of FC layers and LoRA adapters.}
\label{tab:type}
\begin{tabular}{l|l}
\hline \hline
$FC_{y}$	& Compute $\bm{y}$ \\
$FC_{ywbx}$	& Compute $\bm{y}$, $\bm{gW}$, $\bm{gb}$, and $\bm{gx}$ \\
$FC_{ywb}$	& Compute $\bm{y}$, $\bm{gW}$, and $\bm{gb}$ \\
$FC_{ybx}$	& Compute $\bm{y}$, $\bm{gb}$, and $\bm{gx}$ \\
$FC_{yb}$	& Compute $\bm{y}$ and $\bm{gb}$ \\
$FC_{yx}$	& Compute $\bm{y}$ and $\bm{gx}$ \\
\hline
$LoRA_{ywx}$	& Compute $\bm{y_A}$, $\bm{y_B}$, $\bm{gW_B}$, $\bm{gW_A}$, $\bm{gx_B}$, and $\bm{gx_A}$ \\
$LoRA_{yw}$	& Compute $\bm{y_A}$, $\bm{y_B}$, $\bm{gW_B}$, $\bm{gW_A}$, and $\bm{gx_B}$ \\
\hline
\end{tabular}
\end{table}

As basic fine-tuning methods, in this paper, FT-All, FT-Last, and
FT-Bias are defined as follows:
\begin{itemize}
\item FT-All: Weight and bias parameters of all layers are updated.
\item FT-Last: Weight and bias parameters of the last layer are updated.
\item FT-Bias: Bias parameters of all layers are updated.
\end{itemize}
Figures \ref{fig:ft-all}, \ref{fig:ft-last}, and \ref{fig:ft-bias}
illustrate FT-All, FT-Last, and FT-Bias methods, respectively, for
DNNs consisting of three layers.
In these figures, the parameters to be updated are colored in red.
The compute types of the first, second, and third FC layers in FT-All 
are \{$FC_{ywb}$, $FC_{ywbx}$, $FC_{ywbx}$\}.
Those in FT-Last are \{$FC_{y}$, $FC_{y}$, $FC_{ywb}$\}, and those in
FT-Bias are \{$FC_{yb}$, $FC_{ybx}$, $FC_{ybx}$\}.
In the first layer, $\bm{gx}$ is not propagated any more and thus can
be omitted.

%
%
A forward pass of a LoRA adapter computes $\bm{y_A}$ and $\bm{y_B}$,
while the backward pass computes $\bm{gW_B}$, $\bm{gW_A}$,
$\bm{gx_B}$, and $\bm{gx_A}$.
Compute types of LoRA adapters are classified as listed in the lower 
half of Table \ref{tab:type}.
%
%
The compute and memory cost model for each compute type is omitted in
this paper.

%
%

%
%

%
%
As fine-tuning methods of LoRA, LoRA-All and LoRA-Last are defined as
follows:
\begin{itemize}
\item LoRA-All: LoRA adapters are added to all layers.
\item LoRA-Last: A LoRA adapter is added to the last layer.
\end{itemize}
Figures \ref{fig:lora-all} and \ref{fig:lora-last} illustrate LoRA-All
and LoRA-Last methods, respectively, for DNNs consisting of three layers, 
where the parameters to be updated are colored in red.
%
%
The compute types of the first, second, and third LoRA adapters in 
LoRA-All are \{$LoRA_{yw}$, $LoRA_{ywx}$, $LoRA_{ywx}$\}, and those of 
the FC layers are \{$FC_{y}$, $FC_{yx}$, $FC_{yx}$\}.
Similarly, the compute types of the first, second, and third LoRA 
adapters in LoRA-Last are \{$\phi$, $\phi$, $LoRA_{yw}$\}, and those of
the FC layers are \{$FC_{y}$, $FC_{y}$, $FC_{y}$\}.
The backward compute cost of LoRA-Last is thus much smaller than that
of LoRA-All.
On the other hand, LoRA-All introduces LoRA adapters to all the
layers, while LoRA-Last introduces only a single LoRA adapter to the
last layer; thus, LoRA-All has a higher expressive power than LoRA-Last.

\subsection{Performance Analysis}\label{ssec:analysis}

\begin{table}[t]
\centering
\caption{Execution times breakdown of FT-All-LoRA on two datasets.}
\label{tab:time-fc}
\begin{tabular}{l|r|r|l|r|r}
\hline \hline
Forward	&Fan	&HAR	&Backward&Fan	&HAR \\
\hline
FC1	&71.80	&88.58	&FC3	&1.28  	&1.22 \\
LoRA1	&2.75	&1.72	&LoRA3	&1.93  	&1.05 \\
BN1	&2.22	&0.81	&Act2	&0.29  	&0.16 \\
Act1	&0.30	&0.11	&BN2	&2.81  	&1.55 \\
FC2	&17.52	&6.63	&FC2	&34.03 	&18.29 \\
LoRA2	&1.69	&0.61	&LoRA2	&3.30  	&1.78 \\
BN2	&2.23	&0.81	&Act1	&0.29  	&0.15 \\
Act2	&0.30	&0.11	&BN1	&2.84  	&1.53 \\
FC3	&0.50	&0.36	&FC1	&49.47 	&70.46 \\
LoRA3	&0.68	&0.25	&LoRA1	&3.76  	&3.80 \\
\hline
Total (\%)&100.00&100.00&Total (\%)&100.00&100.00\\
\hline
\end{tabular}
\end{table}


%
%
In this section, the compute costs of fine-tuning methods are analyzed.
To analyze the compute costs, ``FT-All-LoRA'' is defined as a full
fine-tuning method that combines FT-All and LoRA-All.
We assume a simple 3-layer DNN that consists of FC layer (FC1), 
LoRA adapter (LoRA1), batch normalization \cite{Ioffe15} (BN1), ReLU (Act1), FC2, 
LoRA2, BN2, Act2, FC3, LoRA3, and cross entropy loss (CEL) function.
Table \ref{tab:time-fc} shows the execution times breakdown of the
forward and backward passes without CEL.
Two datasets, Fan and HAR which will be explained in Section
\ref{ssec:setup}, are examined.
As shown, the first and second FC layers consume most of the compute costs.
To reduce these compute costs, Skip-LoRA and Skip2-LoRA are
proposed in the next section.


\section{Design and Implementation of Skip2-LoRA}\label{sec:design}

\subsection{Proposed Architecture: Skip-LoRA}\label{ssec:skip-lora}

%
%
Our first proposal is ``Skip-LoRA'' which aims to achieve a comparable
expressive power to LoRA-All, yet with a comparable backward compute
cost to LoRA-Last.
Skip-LoRA is defined as follows:
\begin{itemize}
\item Skip-LoRA: LoRA adapters are added between output nodes of the
  last layer and input nodes of the other layers.
\end{itemize}

%
%
As shown in Figures \ref{fig:lora-all} and \ref{fig:lora-last}, weight
parameters of a LoRA adapter for the $k$-th layer are denoted as
$\bm{W^{k-1,k}}$.
For a DNN consisting of $n$ layers, additional weight parameters of 
LoRA-Last and LoRA-All are denoted as $\bm{W^{n-1,n}}$ and 
$\sum^n_{k=1} \bm{W^{k-1,k}}$, respectively.
On the other hand, those of Skip-LoRA are denoted as
$\sum^n_{k=1} \bm{W^{k-1,n}}$.
In this case, the forward pass of all the FC layers is computed 
normally with Equation \ref{eq:fc}.
Then, the forward pass of $n$ LoRA adapters is computed, and the
results are added to the output feature map of the $n$-th FC layer 
as follows:
\begin{equation}
  \bm{y^n} \leftarrow \bm{y^n} + \sum^n_{k=1} \bm{x^k} \cdot \bm{W_A^{k-1,n}} \cdot \bm{W_B^{k-1,n}}, \label{eq:skip2-lora}
\end{equation}
where $\bm{x^k}$ and $\bm{y^k}$ are the input and output feature maps
of the $k$-th FC layer.

%
%
The compute types of the first, second, and third LoRA adapters in 
Skip-LoRA are \{$LoRA_{yw}$, $LoRA_{yw}$, $LoRA_{yw}$\}, and those of 
the FC layers are \{$FC_{y}$, $FC_{y}$, $FC_{y}$\}.
The compute types of FC layers of Skip-LoRA are identical to those 
of LoRA-Last, while the compute types of LoRA adapters are more 
complicated in Skip-LoRA.
Please note that a LoRA adapter is a low-rank approximation of an
FC layer; thus, we can expect $R <<N,M$.
In this case, the computation cost of the FC layers is dominant compared
to that of LoRA adapters, as demonstrated in Table \ref{tab:time-fc}.
We can thus expect that the backward compute cost of Skip-LoRA is
close to that of LoRA-Last while Skip-LoRA has $n$ LoRA adapters to
enhance the expressive power compared to LoRA-Last.

\subsection{Proposed Cache: Skip-Cache}\label{ssec:skip-cache}

%
%
Skip-LoRA can reduce the backward compute cost, as well as LoRA-Last.
The next bottleneck is the forward compute cost.
Here, we aim to reduce the forward compute cost by reusing the forward
compute results which have been already computed.

Let $E$ be the number of fine-tuning epochs.
In the stochastic gradient descent, it is expected that the same
training sample appears $E$ times on average during a fine-tuning process.
Assume we have a set of training samples $\bm{T}$ for the fine-tuning.
Let $\bm{x_i^k} \in \mathbb{R}^{N}$ and $\bm{y_i^k} \in \mathbb{R}^{M}$
be the input and output feature maps of the $k$-th FC layer for the 
$i$-th training sample, where $0 \leq i < |\bm{T}|$.
$\bm{y_i^k}$ is computed and the result is cached as $\bm{c_i^k}$ when
the $i$-th sample appears at the first time, while $\bm{c_i^k}$ is
reused when the $i$-th sample appears again during the fine-tuning
process.
This approach is denoted as ``Skip-Cache'' in this paper.

%
%
Skip-Cache works well if the cached result $\bm{c_i^k}$ is valid
throughout the fine-tuning process over $E$ epochs.
Conversely, Skip-Cache does not work well for FT-All, FT-Bias, and
LoRA-All as illustrated below:
\begin{itemize}
\item FT-All: $\bm{W^k}$ and $\bm{b^k}$ are updated every fine-tuning
  batch, where $1 \leq k \leq n$, obsoleting the cached results frequently.
\item FT-Bias: $\bm{b^k}$ are updated every fine-tuning batch, where
  $1 \leq k \leq n$.
\item LoRA-All: $\bm{W_A^{k-1,k}}$ and $\bm{W_B^{k-1,k}}$ are updated
  every fine-tuning batch, where $1 \leq k \leq n$.
\end{itemize}

%
%
Obviously, Skip-Cache works well for FT-Last, LoRA-Last, and Skip-LoRA,
except for the last FC layer because:
\begin{enumerate}
\item Their output feature maps except for the last layer can be
  computed normally with Equation \ref{eq:fc}, and
\item Their parameters (e.g., $\bm{W^k}$ and $\bm{b^k}$) are not changed
  throughout the fine-tuning process over $E$ epochs, where $1 \leq k < n$.
\end{enumerate}
Please note that a special treatment is needed only for the last layer
(i.e., $k=n$) as illustrated below:
\begin{itemize}
\item FT-Last: The output feature map of the last layer (i.e., $\bm{y_i^n}$)
  cannot be reused because $\bm{W^n}$ and $\bm{b^n}$ are updated every
  fine-tuning batch.
\item LoRA-Last: The result of $G(\bm{x_i^n} \cdot \bm{W^n} + \bm{b^n})$ can be reused as $\bm{c_i^n}$;
  then, $\bm{y_i^n} \gets \bm{c_i^n} + \bm{x_i^n} \cdot \bm{W_A^{n-1,n}} \cdot \bm{W_B^{n-1,n}}$
  is recomputed because $\bm{W_A^{n-1,n}}$ and $\bm{W_B^{n-1,n}}$ are updated
  every fine-tuning batch.
\item Skip-LoRA: $\bm{c_i^n}$ can be reused as well as LoRA-Last; then,
  $\bm{y_i^n} \gets \bm{c_i^n} + \sum_{k=1}^{n} \bm{x_i^k} \cdot \bm{W_A^{k-1,n}} \cdot \bm{W_B^{k-1,n}}$ is
  recomputed because weight parameters of all the $n$ LoRA adapters
  (i.e., $\bm{\forall k,W_A^{k-1,n}}$ and $\bm{\forall k,W_B^{k-1,n}}$ where $1 \leq k \leq n$) are
  updated every fine-tuning batch.
\end{itemize}
In this paper, the combination of Skip-LoRA and Skip-Cache is denoted as
``Skip2-LoRA''.

\subsection{Implementation of Skip2-LoRA}\label{ssec:skip-impl}

\begin{algorithm}[t]
\caption{Fine-tuning with Skip2-LoRA}
\label{alg:skip2-lora}
\begin{algorithmic}[1]
\begin{small}
\Function{ft\_skip2\_lora}{}
\State $\bm{C_{skip}} \gets \phi$ \Comment{Initialize $\bm{C_{skip}}$}
\For {$e = 0$ \textbf{to} $E-1$}
	\For {$b = 0$ \textbf{to} $|\bm{T}|/B-1$}
		\State load\_train\_batch($B$)
		\State forward\_fc($\bm{C_{skip}}$) \Comment{Forward with $\bm{C_{skip}}$}
		\State add\_cache($\bm{C_{skip}}$) \Comment{Add results to $\bm{C_{skip}}$}
		\State forward\_lora()
		\State backward\_lora()
		\State update\_lora\_weight()
	\EndFor
\EndFor
\EndFunction
\end{small}
\end{algorithmic}
\end{algorithm}
\setlength{\textfloatsep}{4mm}

\begin{algorithm}[t]
\caption{FC forward with Skip-Cache}
\label{alg:fc-fw}
\begin{algorithmic}[1]
\begin{small}
\Function{forward\_single\_fc}{$\bm{C_{skip}}$}
\For {$i = 0$ \textbf{to} $B-1$}
	\If {$\bm{x_i} \in \bm{C_{skip}}$} \Comment{If result $\bm{y_i}$ is cached}
		\State \textbf{continue}
	\EndIf
	\For{$m = 0$ \textbf{to} $M-1$}
		\State $y_{i,m} \gets b_m$
		\For {$n = 0$ \textbf{to} $N-1$}
			\State $y_{i,m} \gets y_{i,m} + x_{i,n} \cdot W_{n,m}$ \Comment{Scalar MAC}
		\EndFor
	\EndFor
\EndFor
\State \Return {$\bm{y}$}
\EndFunction
\end{small}
\end{algorithmic}
\end{algorithm}
\setlength{\textfloatsep}{4mm}

%
%
Skip2-LoRA is implemented with the C language without any external
libraries except for libm (``-lm'' option).
Algorithm \ref{alg:skip2-lora} shows the fine-tuning with
Skip2-LoRA algorithm.
$\bm{T}$, $|\bm{T}|$, $E$, and $B$ are the training samples for
fine-tuning, the number of training samples, the number of epochs, and
the batch size, respectively.

In line 2, Skip-Cache $\bm{C_{skip}}$ is initialized.
In line 5, a batch of training samples is randomly selected from $\bm{T}$.
In line 6, the forward pass of all the FC layers is computed for the batch.
During this computation, $\bm{C_{skip}}$ is examined and unnecessary
computation is skipped.
Algorithm \ref{alg:fc-fw} shows the forward pass of a single FC layer with 
$\bm{C_{skip}}$.
This is a typical matrix multiplication algorithm that computes
Equation \ref{eq:fc} except that $\bm{C_{skip}}$ is introduced.
Let $\bm{x_i} \in \mathbb{R}^{N}$ be a training sample in the batch.
If its compute result $\bm{y_i} \in \mathbb{R}^{M}$ has been
cached in $\bm{C_{skip}}$, the computation is skipped so that we can
reduce the compute cost of the forward pass (lines 3-4 of Algorithm 
\ref{alg:fc-fw}).
After completing the forward pass of all the FC layers, newly computed 
results are added to $\bm{C_{skip}}$ as shown in line 7 of 
Algorithm \ref{alg:skip2-lora}.
In line 8, Equation \ref{eq:skip2-lora} is computed.
In lines 9-10, weight parameters of $n$ LoRA adapters are updated.

%
%
As shown in Algorithm \ref{alg:skip2-lora}, $\bm{C_{skip}}$ is
initialized at the beginning of the fine-tuning (line 2), and then the
compute results are continuously added to $\bm{C_{skip}}$ (line 7) 
throughout the fine-tuning process over $E$ epochs.
Since each training sample appears $E$ times on average during a
fine-tuning process, it is expected that the forward compute cost is
reduced to $1/E$.

%
%
Data structure of $\bm{C_{skip}}$ affects the cache hit rate, cache
manipulation overhead, and storage size.
The forward computation for the $i$-th training sample can be skipped
when the compute results $\bm{\forall k,y_i^k}$ where $1 \leq k \leq n$ are
cached in $\bm{C_{skip}}$ \footnote{In reality, the activation function
and batch normalization are typically executed after each layer as in
Table \ref{tab:time-fc}. In this case, the outputs after these
functions should be cached except for the last layer (i.e., the
outputs just after the FC layer should be cached in the case of the
last layer).}.
In the Fan dataset which will be explained in Section \ref{ssec:setup}, 
for example, the number of fine-tuning samples is 470, each containing 
256 features in float32; in this case, the fine-tuning data size is 470KiB.
Assuming a DNN consisting of three layers (e.g., 256-96-96-3), the 
size of $\bm{C_{skip}}$ to store $\bm{\forall i \forall k,y_i^k}$ where 
$1 \leq k \leq n$ is only 358KiB, which is smaller than the input data storage.
Thus, in this paper, $\bm{\forall i \forall k,y_i^k}$ is fully stored in
$\bm{C_{skip}}$.
Since $\bm{\forall k,y_i^k}$ is stored exclusively in the $i$-th element of
$\bm{C_{skip}}$, the time complexity to find the cached results of the
$i$-th training sample is $O(1)$.
Alternatively, if the storage size is strictly limited, a key-value
cache with a limited number of cache entries can be used.
In any cases, there is a trade-off between the cache size and performance.


\section{Evaluations}\label{sec:eval}

The proposed Skip2-LoRA is compared with the counterparts
in terms of accuracy and execution time using drifted datasets.
It is also compared with the state-of-the-art method.

\subsection{Evaluation Setup}\label{ssec:setup}

\begin{figure}[t]
        \centering
        \includegraphics[height=33mm]{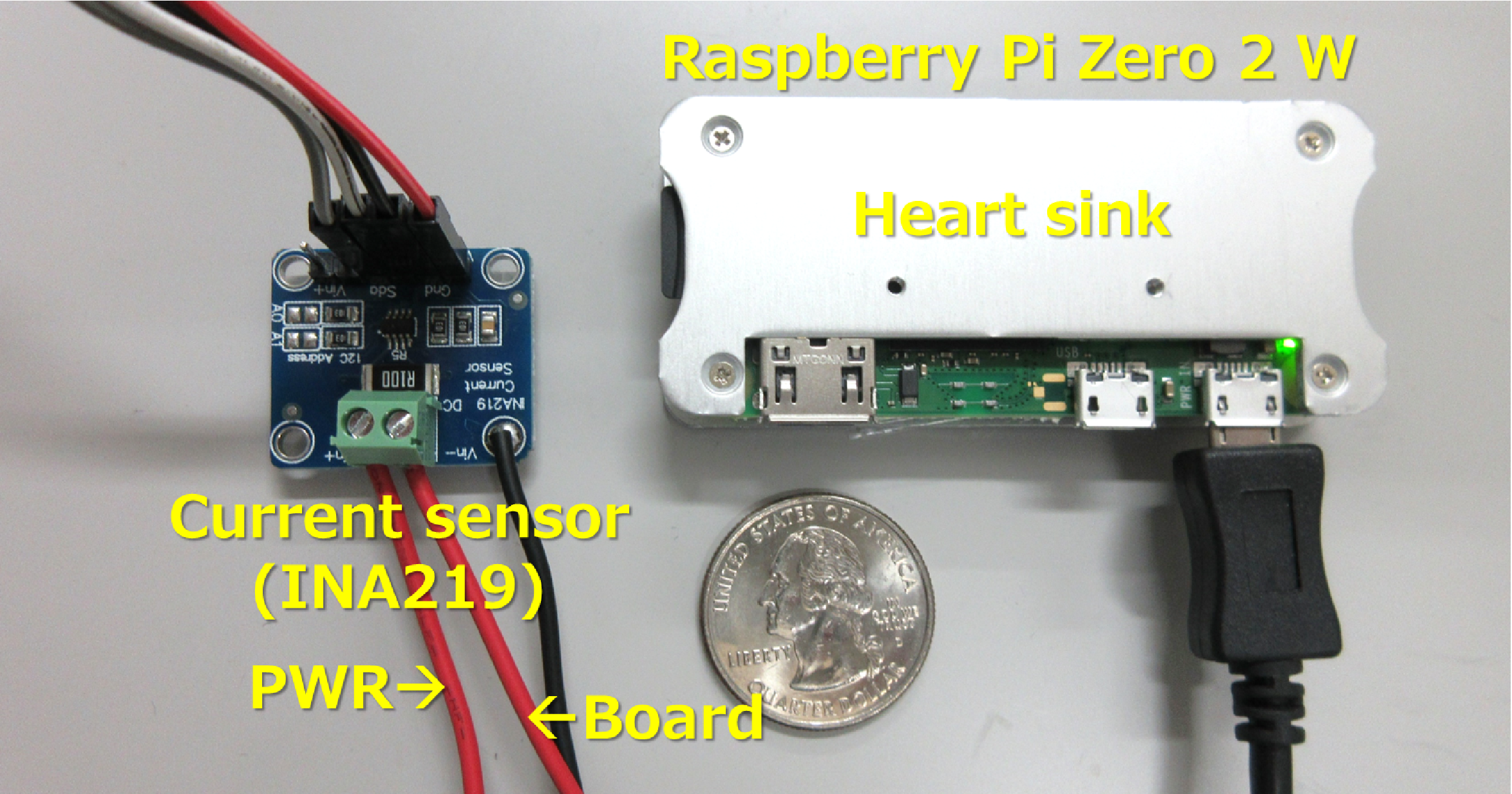}
	\caption{Evaluation environment consisting of Raspberry Pi
          Zero 2 W.}
	\label{fig:rasp-zero2}
\end{figure}

%
%
FT-All, FT-Last, FT-Bias, FT-All-LoRA, LoRA-All, LoRA-Last, Skip-LoRA,
and Skip2-LoRA are executed on a Raspberry Pi Zero 2 W board \cite{RPiZero2}, 
which is known as a \$15 computer (Figure \ref{fig:rasp-zero2}).
The clock frequency is fixed at 1GHz to measure the execution time stably.
Skip2-LoRA and its counterparts (except for TinyTL \cite{Cai20}) are 
implemented with the C language and compiled with gcc version 8.3.0
with ``-O3'' option on the platform.
They are further optimized with SIMD (Neon) instructions with
``-mfpu=neon -ffast-math'' option.

\begin{table}[t]
\centering
\caption{Accuracy of before and after data drift on 3-layer DNN (\%).}
\label{tab:acc-base}
\begin{tabular}{l|r|r}
\hline \hline
	&Before		&After		\\
\hline
Damage1	&60.61$\pm$13.73&98.99$\pm$2.81	\\
Damage2	&51.86$\pm$8.04	&90.88$\pm$5.65	\\
HAR	&79.97$\pm$5.62	&86.09$\pm$4.40	\\
\hline
\end{tabular}
\end{table}

\begin{table*}[t]
\caption{Accuracy of proposed and counterpart fine-tuning methods on
  3-layer DNN (\%).}
\hspace{-12mm}
\label{tab:acc}
\begin{tabular}{l|r|r|r|r|r|r|r|r}
\hline \hline
	&FT-All		&FT-Last	&FT-Bias	&FT-All-LoRA	&LoRA-All	&LoRA-Last	&Skip-LoRA	&Skip2-LoRA	\\
\hline
Damage1	&98.73$\pm$2.11	&94.19$\pm$2.24	&79.42$\pm$7.50	&98.63$\pm$2.14	&98.26$\pm$1.32	&94.67$\pm$2.92	&96.07$\pm$2.14	&96.19$\pm$2.29	\\
Damage2	&88.12$\pm$6.13	&92.43$\pm$3.67	&79.56$\pm$6.47	&88.88$\pm$5.73	&86.45$\pm$4.90	&93.55$\pm$3.50	&93.24$\pm$3.86	&93.46$\pm$3.21	\\
HAR	&90.99$\pm$1.86	&89.31$\pm$1.06	&82.21$\pm$1.27	&90.40$\pm$2.49	&91.09$\pm$1.26	&89.79$\pm$1.46	&92.10$\pm$1.05	&91.99$\pm$1.00	\\
\hline
\end{tabular}
\end{table*}

%
%
To evaluate the fine-tuning methods, we use three datasets each
containing pre-train samples, fine-tune samples, and test samples as follows:
\begin{itemize}
\item Damage1 is a 3-class classification task of vibration patterns of cooling
  fans (i.e., stop, normal fan, and damaged fan with holes on a blade).
  Both the normal and damaged fans rotate at 1,500, 2,000, and 2,500 rpm.
  The numbers of input features and output classes are 256 and 3.
  The original dataset \cite{Sunaga23} contains vibration patterns in
  a silent office and those near a ventilation fan (they are denoted
  as ``silent dataset'' and ``noisy dataset'').
  The model is pre-trained with the silent dataset.
  We assume the model is deployed in a ``real'' noisy environment.
  Thus, the model is fine-tuned with a half of the noisy dataset and
  then tested with the remaining half of the noisy dataset.
  The numbers of pre-train, fine-tune, and test samples are 470, 470, and 470.
\item Damage2 is similar to the Damage1 dataset, but using a damaged fan
  with a chipped blade.
\item HAR is a 6-class classification task of human activity recognition (i.e.,
  walking, walking upstairs, walking downstairs, sitting, standing,
  and laying).
  The numbers of input features and output classes are 561 and 6.
  The original dataset \cite{Reyes12} contains sensor data from 30
  human subjects.
  We manually removed those of subjects 9, 14, 16, 19, and 25 from the
  original dataset and saved as ``initial dataset''.
  Those of subjects 9, 14, 16, 19, and 25 were saved as ``drifted dataset''.
  The model is pre-trained with the initial dataset.
  We assume the model is used for a group of different subjects.
  Thus, the model is fine-tuned with a half of the drifted dataset and
  then tested with the remaining half of the drifted dataset.
  The numbers of pre-train, fine-tune, and test samples are 5,894, 1,050,
  and 694.
\end{itemize}

%
%
We use a simple 3-layer DNN shown in Figure \ref{fig:ft-method}.
The numbers of input and output nodes are 256 and 3 for the Damage1 and Damage2
datasets.
They are 561 and 6 for the HAR dataset.
The number of hidden nodes is 96 in all the hidden layers.
The LoRA rank is set to 4.
Batch normalization and ReLU are also executed as in Table \ref{tab:time-fc}.

\subsection{Accuracy}\label{ssec:accuracy}

%
%
Table \ref{tab:acc-base} shows the baseline accuracy of the three
datasets without fine-tuning on the 3-layer DNNs.
In the ``Before'' case, the model is trained with the pre-train
dataset and then tested with the test dataset.
In the ``After'' case, the model is trained only with the fine-tune
dataset and then tested with the test dataset.
In each case, the number of training epochs is set to a large enough
value (i.e., $E=400$ and $900$ for the Damage1/Damage2 and HAR datasets).
Table \ref{tab:acc-base} shows mean accuracy values over 20 trials.
The accuracy is quite low in the Before case while it is significantly
better in the After case.
There is a significant accuracy gap between before and after the data
drift; in this case, we can fill out the gap by the on-device
fine-tuning as demonstrated below.

%
%
Table \ref{tab:acc} shows the accuracies of FT-All, FT-Last, FT-Bias, 
FT-All-LoRA, LoRA-All, Skip-LoRA, and Skip2-LoRA on the 3-layer DNNs.
The test is conducted with the following three steps.
In each case, the number of training epochs is set to a large enough value.
\begin{enumerate}
\item The model is trained with the pre-train samples
  ($E=100$ and $300$ for the Damage1/Damage2 and HAR datasets).
\item The model is fine-tuned with the fine-tune samples
  ($E=300$ and $600$ for the Damage1/Damage2 and HAR datasets).
\item The model is tested with the test samples.
\end{enumerate}

\begin{table}[t]
\centering
\caption{Accuracy of state-of-the-art fine-tuning methods \cite{Cai20}
	on ProxylessNAS \cite{Cai19} (\%).}
\label{tab:acc-tl}
\begin{tabular}{l|r|r}
\hline \hline
	& TinyTL (GN)	& TinyTL (BN) \\
\hline
Damage1	&98.66$\pm$0.76	&99.49$\pm$0.32 \\
Damage2	&92.09$\pm$3.17	&96.01$\pm$2.74 \\
HAR	&88.76$\pm$0.91	&89.27$\pm$1.13 \\
\hline
\end{tabular}
\end{table}

%
%
Mean accuracies over 20 trials are reported in this table.
In all cases, Skip2-LoRA shows almost the same accuracy as Skip-LoRA.
Note the difference between FT-All and the ``After'' case in Table
\ref{tab:acc-base} is that FT-All is trained with both the pre-train and
fine-tune datasets.
For the Damage1 dataset, Skip2-LoRA achieves a higher accuracy than
FT-Last, FT-Bias, and LoRA-Last, demonstrating a higher expressive
power than these counterparts.
However, its accuracy is lower than FT-All, FT-All-LoRA, and LoRA-All.
For the Damage2 dataset, on the other hand, Skip2-LoRA shows a higher
accuracy than FT-All, FT-All-LoRA, and LoRA-All.
We expect that these counterparts cause overfitting to the ``after drift''
fine-tune dataset and thus they show a lower accuracy than
Skip2-LoRA.
For the HAR dataset, Skip-LoRA and Skip2-LoRA show the highest
accuracies followed by LoRA-All, FT-All, FT-All-LoRA, LoRA-Last,
FT-Last, and FT-Bias.

%
%
Table \ref{tab:acc-tl} shows the accuracies of TinyTL \cite{Cai20}
as a state-of-the-art fine-tuning method, where
``GN'' and ``BN'' mean the group normalization \cite{YWu18} and
batch normalization.
TinyTL uses GN \cite{Cai20}, while its BN version is also tested.
The number of trials is 20 in each case.
The accuracy of Skip2-LoRA is not higher than that of TinyTL in the
Damage1 dataset, while Skip2-LoRA outperforms TinyTL in the HAR
dataset.
Please note that the backbone network of TinyTL is ProxylessNAS
\cite{Cai19} while ours use much simpler 3-layer DNNs.

\subsection{Execution Time}\label{ssec:exectime}



\begin{table*}[t]
\caption{Execution time on Raspberry Pi Zero 2 W with Neon
  instructions for Fan dataset (msec).}
\hspace{-5mm}
\label{tab:speed_fan_neon}
\begin{tabular}{l|r|r|r|r|r|r|r|r}
\hline \hline
		&FT-All	&FT-Last&FT-Bias&FT-All-LoRA	&LoRA-All	&LoRA-Last	&Skip-LoRA	&Skip2-LoRA \\
\hline
Train@batch	&5.864	&2.633	&3.721	&6.053		&4.113		&2.642		&2.952		&{\bf 0.450} \\
~~~forward	&2.812	&2.601	&2.832	&2.868		&2.942		&2.613		&2.807		&0.309 \\
~~~backward	&2.866	&0.030	&0.885	&2.993		&1.157		&0.026		&0.136		&0.131 \\
~~~weight update&0.186	&0.002	&0.003	&0.192		&0.014		&0.002		&0.010		&0.010 \\
\hline
Predict@sample	&0.142	&0.144	&0.148	&0.150		&0.155		&0.143		&0.151		&0.154 \\
\hline
\end{tabular}
\end{table*}

\begin{table*}[t]
\caption{Execution time on Raspberry Pi Zero 2 W with Neon
  instructions for HAR dataset (msec).}
\hspace{-5mm}
\label{tab:speed_har_neon}
\begin{tabular}{l|r|r|r|r|r|r|r|r}
\hline \hline
		&FT-All	&FT-Last&FT-Bias&FT-All-LoRA	&LoRA-All	&LoRA-Last	&Skip-LoRA	&Skip2-LoRA \\
\hline
Train@batch	&11.323	&6.179	&6.795	&11.577		&7.459		&6.031		&6.328		&{\bf 0.595} \\
~~~forward	&6.569	&6.129	&6.050	&6.660		&6.390		&6.005		&6.130		&0.396 \\
~~~backward	&4.373	&0.047	&0.742	&4.480		&1.052		&0.024		&0.184		&0.185 \\
~~~weight update&0.381	&0.003	&0.003	&0.437		&0.017		&0.002		&0.014		&0.014 \\
\hline
Predict@sample	&0.308	&0.307	&0.304	&0.317		&0.314		&0.309		&0.314		&0.317 \\
\hline
\end{tabular}
\end{table*}

\begin{figure*}[t]
\begin{minipage}[t]{0.33\linewidth}
        \centering
        \subfigure[Damage1] {
        \includegraphics[height=36mm]{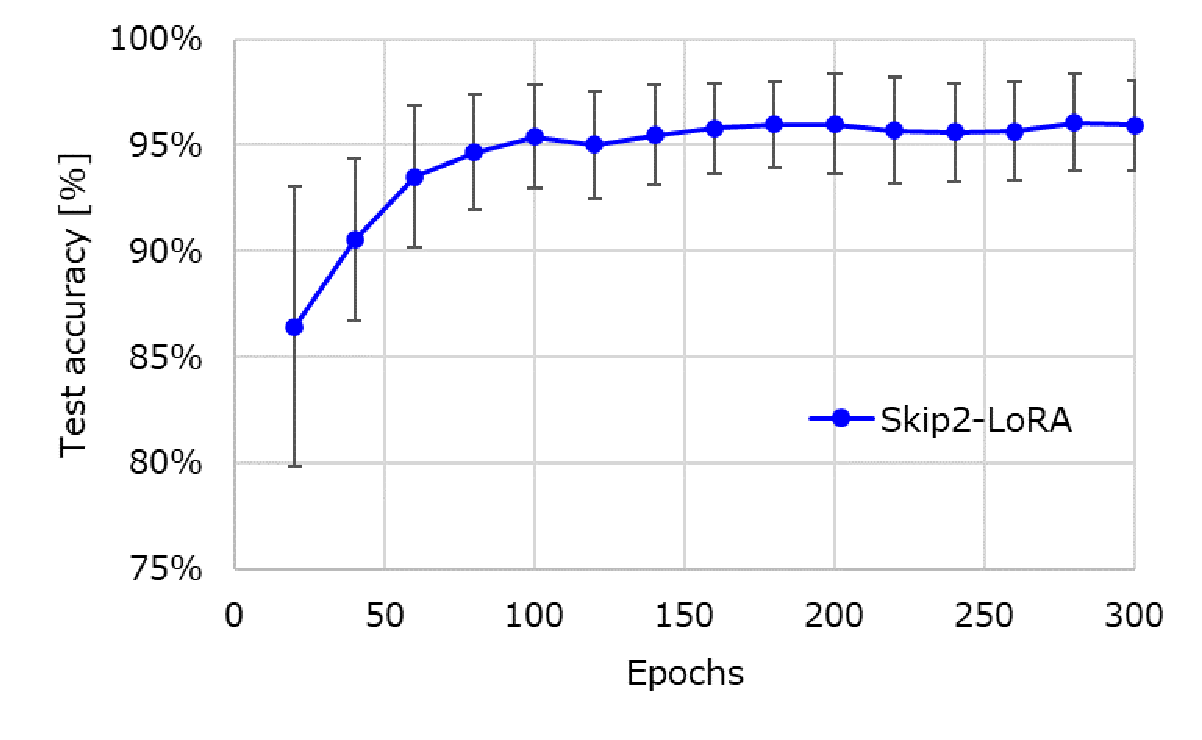}
        \label{fig:curve-damage1}
        }
\end{minipage}
\begin{minipage}[t]{0.33\linewidth}
        \centering
        \subfigure[Damage2] {
        \includegraphics[height=36mm]{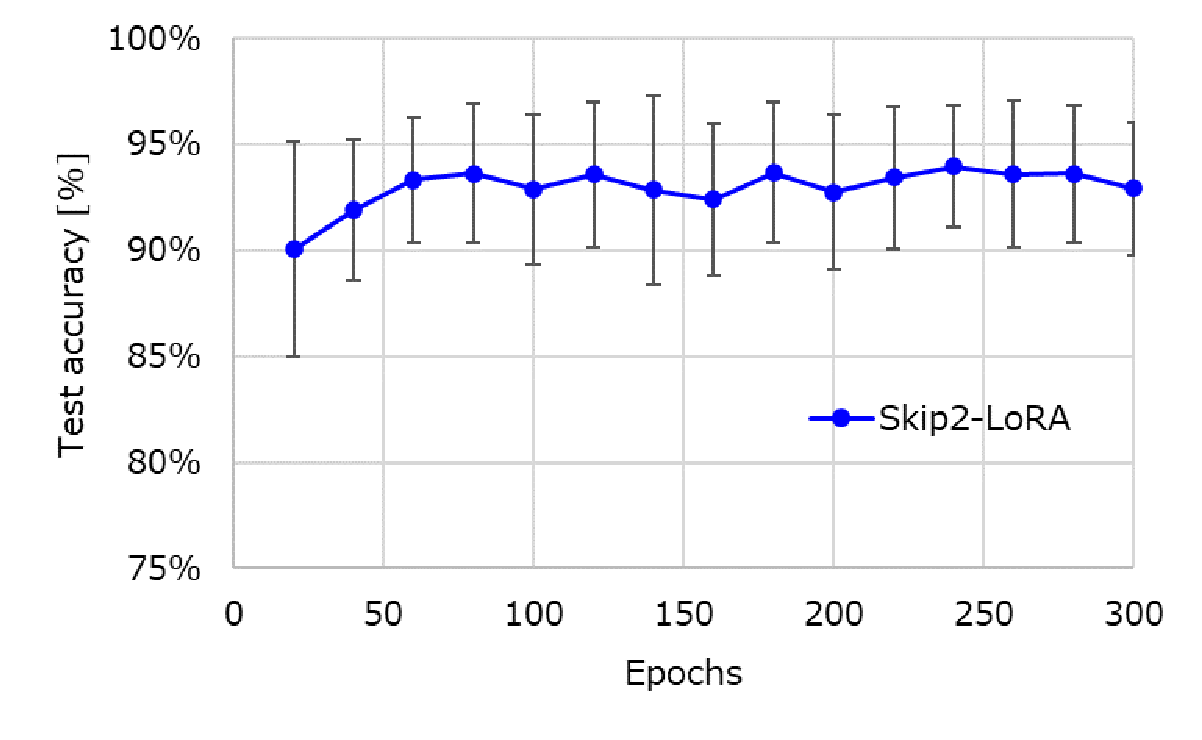}
        \label{fig:curve-damage2}
        }
\end{minipage}
\begin{minipage}[t]{0.33\linewidth}
        \centering
        \subfigure[HAR] {
        \includegraphics[height=36mm]{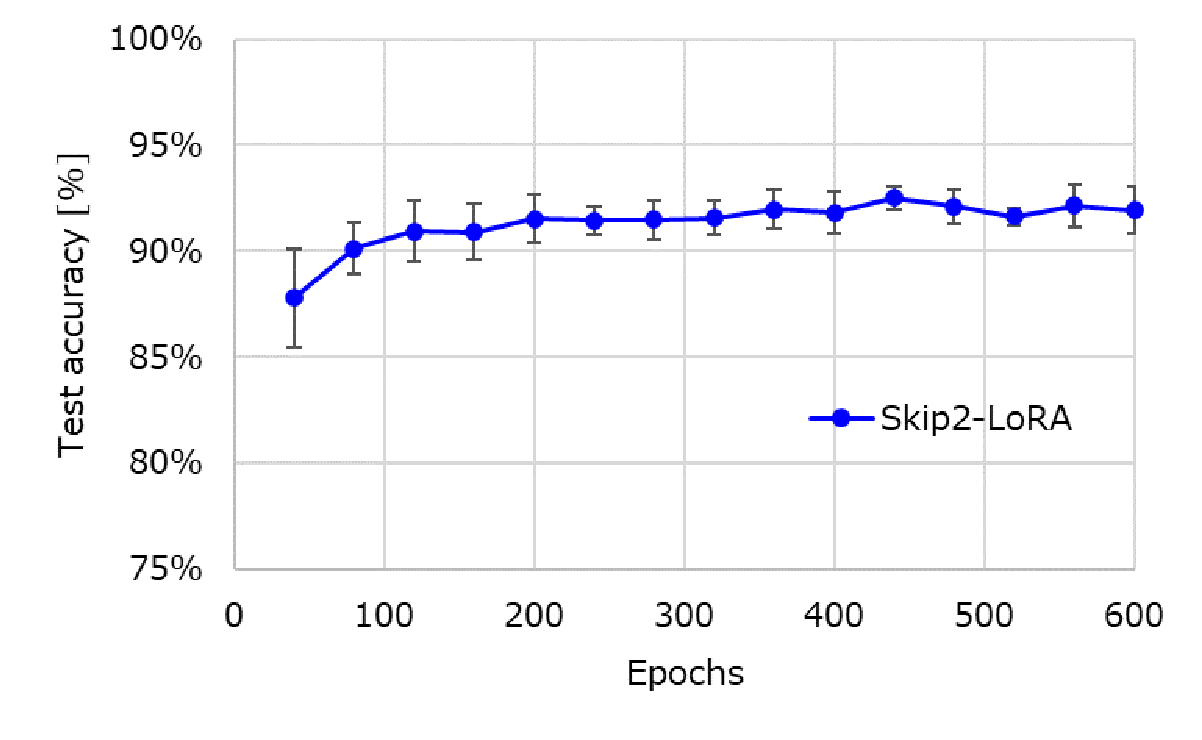}
        \label{fig:curve-har}
        }
\end{minipage}
\caption{Training curves of Skip2-LoRA on three datasets. Required epochs
	are 100, 60, and 200 in Damage1, Damage2, and HAR datasets.}
\label{fig:curve}
\end{figure*}

\begin{figure}[t]
        \centering
        \includegraphics[height=42mm]{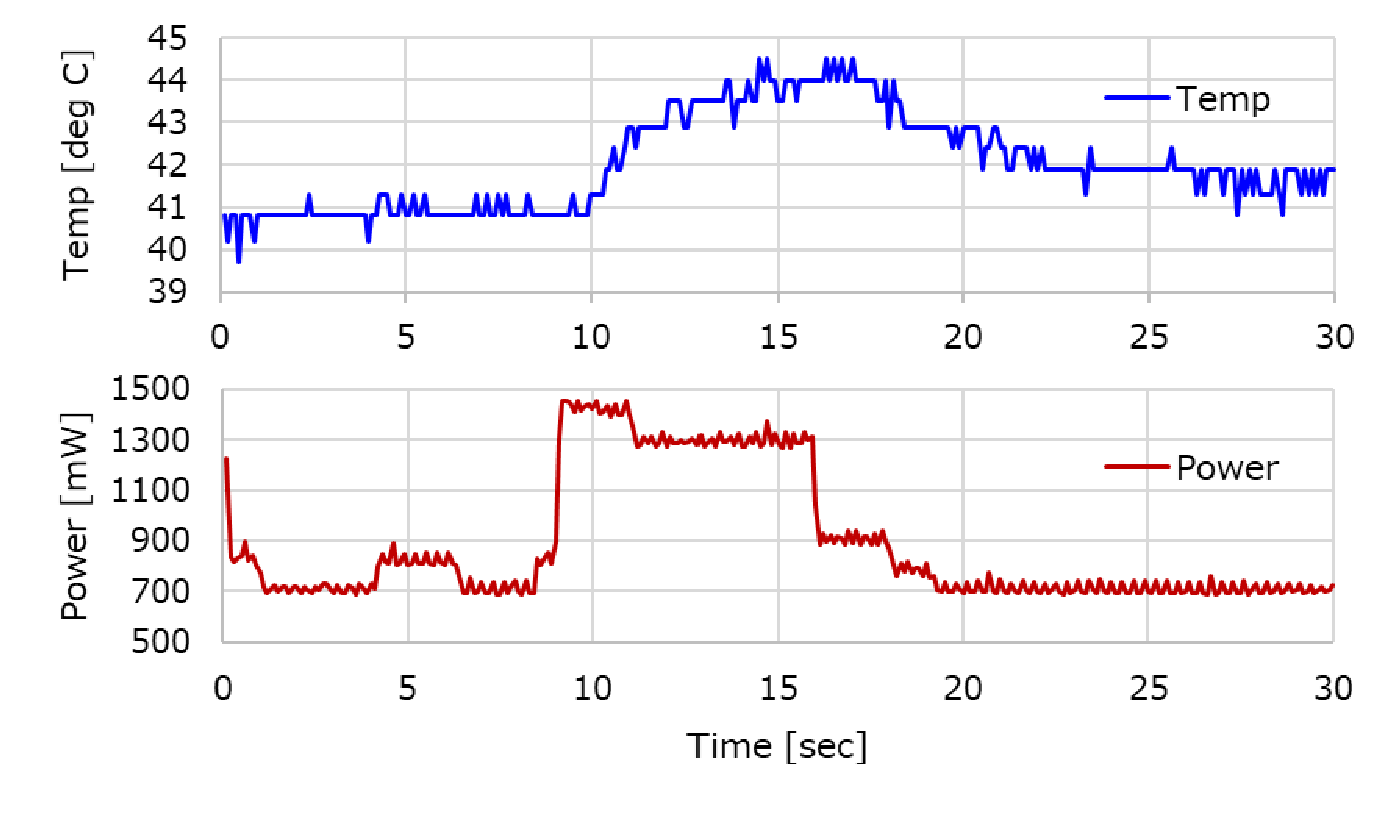}
	\caption{Power consumption and temperature of Skip2-LoRA
          with HAR dataset. Fine-tuning starts at 9sec.}
	\label{fig:power}
\end{figure}

%
%
Tables \ref{tab:speed_fan_neon} and \ref{tab:speed_har_neon} show the
execution times of FT-All, FT-Last, FT-Bias, FT-All-LoRA, LoRA-All,
LoRA-Last, Skip-LoRA, and Skip2-LoRA on a Raspberry Pi Zero 2 W with
Neon instructions.
The results on the Damage1 and Damage2 datasets are the same and thus reported
as ``Fan'' dataset in Table \ref{tab:speed_fan_neon}.
The training batch size $B$ is set to 20.

As shown in these tables, the training time of a batch consists of the
forward pass, backward pass, and weight update.
The execution times are mean values over the entire fine-tuning
process, where the number of epochs $E$ is the same as that in Section
\ref{ssec:accuracy}.
Although Skip-LoRA and LoRA-All have the same number of trainable
parameters, Skip-LoRA reduces the execution time of backward pass by
82.5\% to 88.3\% compared to LoRA-All, demonstrating benefits of the
proposed Skip-LoRA architecture.
In addition, Skip2-LoRA reduces the execution time of forward pass
by 89.0\% to 93.5\% compared to Skip-LoRA, demonstrating benefits of
the proposed Skip-Cache.
As a result, Skip2-LoRA reduces the training time by 89.0\% to 92.0\% 
(90.0\% on average) compared to LoRA-All that has the same number of trainable
parameters.
The training times are only 0.450msec and 0.595msec per
batch in the Fan and HAR datasets, respectively.

%
%
In Skip2-LoRA, the training time is affected the number of epochs $E$,
because a larger $E$ can skip more forward pass computations.
As mentioned in Section \ref{ssec:accuracy}, $E$ was set to a large
enough value.
Here, we estimate actual training time based on practical $E$.
Figure \ref{fig:curve} shows the training curves of Skip2-LoRA with the 
three datasets.
X-axis shows the number of trained epochs, and Y-axis shows the test
accuracy.
Mean accuracies over 10 trials are reported in these graphs.
Here, the number of required epochs in which the test accuracy first
reaches within a 1\% range of the reported accuracies in Table
\ref{tab:acc} is denoted as ``required epochs''.
The required number of epochs are 100, 60, and 200 in the Damage1,
Damage2, and HAR datasets, respectively.
The number of their fine-tuning samples are 470, 470, and 1050;
thus, the total fine-tuning times of Skip2-LoRA on a
Raspberry Pi Zero 2 W are only 1.06sec, 0.64sec, and 2.79sec in the Damage1,
Damage2, and HAR datasets, respectively.



\subsection{Power Consumption}\label{ssec:power}

%
%
Skip2-LoRA with the HAR dataset ($E=200$) is executed on a Raspberry Pi
Zero 2 W and the power consumption is measured with a current sensor
INA219 (Figure \ref{fig:rasp-zero2}).
Figure \ref{fig:power} shows the variation of power consumption and
temperature, where the fine-tuning starts at 9sec.
Once the fine-tuning starts, the clock frequency increases from 600MHz
to 1GHz and the power consumption increases.
Although the net compute time for the forward and backward passes is
2.79sec as mentioned above, the results in Figure \ref{fig:power}
include overheads for reading the dataset and loading the pre-trained
weight parameters.
The power consumption is at most 1,455mW for a short duration and the
temperature does not exceed 44.5 deg C.


\section{Summary}\label{sec:conc}

In this paper, we extended LoRA adapters as a new lightweight on-device
fine-tuning mehtod for resource-limited edge devices.
The proposed Skip2-LoRA synergistically combines Skip-LoRA
architecture to reduce the backward compute cost and Skip-Cache to
reduce the forward compute cost.
Experimental results using three drifted datasets demonstrated that
Skip2-LoRA reduces the fine-tuning time by 90.0\% on average compared
to LoRA-All that has the same number of trainable parameters while
achieving comparable accuracies to the state-of-the-art method.
The order of magnitude reduction of the compute cost enables a few
seconds ``quick'' fine-tuning of DNNs on a Raspberry Pi Zero 2 W
board with modest power and temperature.


{\bf Acknowledgements~~}
H.M. was supported in part by JST AIP Acceleration Research JPMJCR23U3, Japan.


\end{document}